\documentclass[journal]{IEEEtran}

\usepackage{mathrsfs}

\usepackage{graphicx}
\usepackage{amsmath}
\usepackage{multirow}
\usepackage{mathtools}
\usepackage{placeins}
\usepackage{float}

\usepackage{booktabs}
\usepackage{ragged2e}
\usepackage{algorithm}
\usepackage{algorithmic}
\usepackage{amsmath}
\usepackage{graphics}
\usepackage{epsfig}
\usepackage{amssymb}
\usepackage{threeparttable}
\usepackage{stmaryrd}
\usepackage{romannum}

\usepackage{bm}

\ifCLASSINFOpdf
\else
\fi

\begin{document}

\title{Modeling sequential annotations for sequence labeling with crowds}

\author{Xiaolei~Lu, Tommy~W.S.Chow,~\IEEEmembership{Fellow,~IEEE}%

\IEEEcompsocitemizethanks{\IEEEcompsocthanksitem Xiaolei Lu is with
the Dept of Electrical Engineering at the City University of Hong
Kong, Hong Kong (Email:xiaoleilu2-c@my.cityu.edu.hk)
\IEEEcompsocthanksitem Tommy W S Chow is with the Dept of Electrical
Engineering at the City University of Hong Kong, Hong Kong (E-mail:
eetchow@cityu.edu.hk).}}

\maketitle

\begin{abstract}
Crowd sequential annotations can be an efficient and cost-effective way to build large datasets for sequence labeling. Different from tagging independent instances, for crowd sequential annotations the quality of label sequence relies on the expertise level of annotators in capturing internal dependencies for each token in the sequence. In this paper, we propose Modeling sequential annotation for sequence labeling with crowds (SA-SLC). First, a conditional probabilistic model is developed to jointly model sequential data and annotators' expertise, in which categorical distribution is introduced to estimate the reliability of each annotator in capturing local and non-local label dependency for sequential annotation. To accelerate the marginalization of the proposed model, a valid label sequence inference (VLSE) method is proposed to derive the valid ground-truth label sequences from crowd sequential annotations. VLSE derives possible ground-truth labels from the token-wise level and further prunes sub-paths in the forward inference for label sequence decoding. VLSE reduces the number of candidate label sequences and improves the quality of possible ground-truth label sequences. The experimental results on several sequence labeling tasks of Natural Language Processing show the effectiveness of the proposed model.

\end{abstract}


\section{Introduction}

\IEEEPARstart{S}{equence} labeling, has been widely applied in Natural Language Processing and Computational Biology. Given a sequence of observations, sequence labeling assigns an appropriate label for each observation. For example, Part-of-Speech (POS) tagging can be formulated as a sequence labeling task, which predicts the POS of each word for the sentence sequence and produces a POS sequence of equal length.

There have been many statistical models proposed with promising results for sequence labeling (e.g. structured SVMs (S-SVM) \cite{re1} and graphical models). Similar to other supervised learning methods, sequence labeling requires large number of training sequences with complete annotations, which is costly and laborious to produce. As semi-supervised learning can perform learning task from a small amount of labeled data and large amounts of unlabeled data, semi-supervised sequence labeling has exhibited advantages over full manual annotation. However, semi-supervised based methods still need exact annotations.

In recent years crowdsourcing has been widely adopted to obtain large labeled datasets in a cheap and efficient way. Crowdsourcing is the process of obtaining a large amount of labeled data from a group of ordinary people. It does not require domain experts but relies on utilizing the contribution of the group's intelligence \cite{re2}. But the quality of labels obtained from crowdsourcing cannot be guaranteed because the expertise level of annotators varies. Therefore learning from crowd labels still mainly focuses on estimating the reliability of individual annotator and building classifiers to predict new input data in aggregation process. For example, Snow et al. \cite{re3} used bias correction to combine non-expert annotation. Raykar et al. \cite{re4} proposed to jointly estimate the coefficients of a logistic regression classifier and the annotators’ expertise. Yan et al. \cite{re5} further considered the case that the reliability of an annotator is not consistent across all the input data.

Crowdsourcing has been exploited in sequence labeling. Early work extended the methods (e.g. majority voting and weighted voting \cite{re6}) developed for independent instances with crowds to handle crowd sequential annotations, which aims to infer the ground-truth label for each token of the sequence. To better model crowd sequential annotation, studies on traditional sequence labeling models (e.g. HMMs \cite{re7} and CRFs \cite{re8}) have been conducted aiming to include additional modeling of the quality of crowd label sequences. These methods generally treat each token as an independent instance and assume that the label quality for each token by annotators is only a function of their level of expertise. However, there exists internal label dependency among tokens in a sequence, which is different from instances that are independent from each other. For example, given the sentence ``This inconsistency with observation sent Albert Einstein back to the drawing board and, on 25 November 1915, Einstein presented the updated field equations to the Prussian Academy of Sciences" for named entity recognition, the label dependency constrains all the mentions of ``Einstein" located in different tokens to share the same label with ``PERSON" \cite{re9}. Therefore, for crowd sequential annotations the quality of label sequence relies on annotators' expertise in capturing internal dependencies for each token in a given sequence. 

In this paper, to improve the estimation of the quality of label sequence of crowd sequential annotations, we propose a modeling sequential annotation method for sequence labeling with crowds (SA-SLC). Through modeling annotator's expertise in capturing internal label dependency for sequential annotation, SA-SLC enhances the quality of measuring the whole label sequence and then assigns different weights to crowd label sequences for classifier learning. As a result, the negative effect of unreliable annotations in the optimization can be significantly reduced. Our contributions can be summarized as follows: 

First, a conditional probabilistic model is developed to jointly model sequential data and annotators' reliability. By decomposing internal label dependency into local and non-local label dependency, we introduce categorical distribution to estimate annotators' expertise in capturing internal label dependency, which results in improving estimation of the quality of label sequences in aggregation process. The concept of local and non-local label dependency will be elaborated in later section of the paper. Expectation maximization (EM) is employed for parameter estimation, where in $M$ step a weighted sequential model is defined to learn parameters for predicting new sequences.

Second, to accelerate the marginalization of the proposed model, valid label sequence inference (VLSE) is proposed to derive the valid ground-truth label sequences from crowd sequential annotations. VLSE first estimates the possible ground-truth labels in the token-wise level and prunes certain sub-paths that violate label constraint in the forward inference for label sequence decoding. As a result, the number of candidate label sequences for marginalization is substantially reduced and the efficiency of parameter estimation can be improved.

To demonstrate the effectiveness of the proposed model, we conduct experiments on the datasets collected from news articles and biomedical papers. The results show that our proposed model performs stably with varying annotators' expertise and in most cases achieves better performance than the state-of-the-art methods.

\section {Related work}

There are two main types of approaches for sequence labeling: generative and discriminative. Hidden Markov Models (HMMs) \cite{re10,re11} and Conditional Random Fields (CRFs) \cite{re12} form the most popular generative-discriminative pair for sequence labeling. Many variants derived from these basic graphical models have been applied to handle complex sequence labeling tasks. For example, Coupled HMM \cite{re13} is developed to model the interactions between two state (i.e. label) sequences to capture more information from the structure. Similarly, hierarchical conditional random fields \cite{re14} model multiple output sequences for joint sequence labeling tasks (e.g. POS tagging and Chunking). Furthermore, Sarawagi and Cohen \cite{re15} proposed semi-Markov conditional random fields (semi-CRFs) to predict and segment the output sequence, which is more appropriate for the tasks of chunking and named entity recognition compared with CRFs. In recent years, the models combining deep learning and graphical models achieve competitive results compared with traditional models, such as CRF-CNN \cite{re16} and Bi-LSTM-CRF \cite{re17}.

Early work on sequence labeling has obtained promising results. But collection of large number of training datasets with exact annotations are still a highly challenging issue. First, it is financially expensive to employ the annotators with high level of expertise who can provide precise labelings. Further, it is prohibitively time-consuming to manually label complex structured outputs than traditional discrete values. For example, annotation for named entity recognition and POS tagging require labeling each word in all sentences level.

Semi-supervised sequence labeling has received increasing attention as it requires only small amount of labeled data. Semi-supervised CRFs \cite{re18} minimizes the conditional entropy on unlabeled training instances and then combines with the objective of CRFs, which improves the performance of supervised CRFs. Brefeld and Scheffer \cite{re19} proposed to incorporate co-training principle into support vector machine to minimize the number of errors for labeled data and the disagreement for the unlabeled data, which can outperform fully-supervised SVM in specific tasks. It is worth noting that semi-supervised sequence labeling partly lighten the burden of sequential annotations. But semi-supervised sequence labeling still needs exact labelings.

Crowdsourcing provides an efficient and cost-effective way to obtain large amounts of labeled datasets. However, the label quality cannot be guaranteed as the expert level of annotators varies. Traditional way is to infer the ground-truth label from crowd annotations. For example, majority voting \cite{re20} and weighted voting \cite{re6}. Since the subsequent classifier learning heavily depends on the inferred ground-truth label, Raykar et al. \cite{re4} developed a probabilistic framework to jointly learn the classifier and the ground-truths, where annotators' expertise and the ground-truths are estimated iteratively. To improve this probabilistic framework, Raykar et al. \cite{re21} used spammer score to rank annotators, which aims to eliminate the spammers and infer the ground-truths based on the reliable annotators in each iteration. Yan et al. \cite{re5} redefined annotator's expertise by assuming that the reliability of annotators is not consistent across all the input data. 

The above crowdsourcing learning models perform on single-instance level. There have been research work studying crowd sequential annotations. HMM-Crowd \cite{re22} is developed to aggregate crowd sequential annotations in the Hidden Markov Model to infer a best single sequential annotation, while BSC \cite{re23} is a fully Bayesian approach that further models the effect of local sequential dependencies on annotators' reliability. Wu et al. \cite{re24} extended the probabilistic framework of Raykar et al. \cite{re4} to crowd sequential labeling by aggregating multiple annotations with weighted CRF model. Rodrigues et al. \cite{re20} proposed CRF-MA that assumes the reliability of an annotator is consistent among all the labels, which models the reliability of an annotator as a multinomial random variable and then treats it as a latent variable in CRF model. 

It is worth noting that sequential annotation aims to provide a reasonable label sequence for the whole input sequence. Unlike the instances that are independent from each other, there exists internal label dependency among the tokens in the sequence. In this paper, different from the above crowdsourcing sequential models, the proposed model measures annotators' expertise and view in capturing internal label dependency among the sequence, which aims to improve the estimation of the quality of label sequence from crowd sequential annotations and enhance the performance of predicting new sequences as a result.

\section {Preliminaries}

\subsection {Learning from crowd labels}
Given the dataset $X = \left \{ x_{i} \right \}_{i=1}^{N}$, let $Y = \left \{ y_{i}^{1}, y_{i}^{2},..., y_{i}^{K } \right \}_{i=1}^{N}$  where $\left \{ y_{i}^{1}, y_{i}^{2},..., y_{i}^{K } \right \}$ denote the  crowd labels for the $i_{th}$ data by $K$ different annotators. $Z=\left \{ z_{i} \right \}_{i=1}^{N}$ represents the ground-truth label set which is unknown to the annotators. Given the training set with crowd labels, the goal of modeling is to estimate the annotators' expertise and a classifier to predict labels for new instances. 

Formulating a probabilistic model of the learning process is a principled way to explore crowd labels. Based on the graphical model as shown in Figure 1, the joint conditional distribution can be defined as 
\begin{equation}
p(Y,Z|X)=\prod_{i}p(z_{i}|x_{i})\prod_{k}p(y_{i}^{k}|x_{i},z_{i}),
\end{equation}
where $p(y_{i}^{k}|x_{i},z_{i})$ measures the annotators' expert level which in most cases is expressed as
\begin{equation}
 \begin{aligned}
 p(y_{i}^{k}|x_{i},z_{i}) &=p(y_{i}^{k}|z_{i})\\
 &= (1-\eta ^{k})^{\left | y_{i}^{t} -z_{i}\right |}\eta ^{k^{1-{\left | y_{i}^{t} -z_{i}\right |}}},\\
 \end{aligned}
\end{equation}
where $\eta^{k}$ denotes the expert level of $k_{th}$ annotator.

\begin{figure}[H]
\centering
\includegraphics[width=1in,height = 0.8 in ]{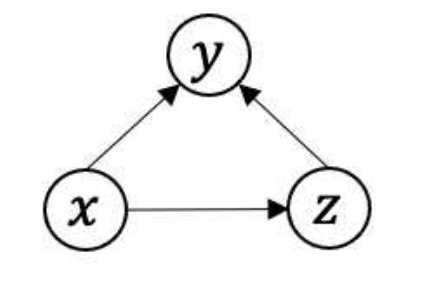}
\caption {Graphical model for $x$, $y$ and $z$.}
\label{fig:secondfigure}
\end{figure}

Generally, for the independent instance with crowd labels, it is reasonable to assume $y_{i}^{k}$ rely on $z_i$ only as no internal annotation dependency need to be considered. 

\subsection {Sequence labeling}
Sequence labeling, assigns a label sequence to the sequential input data. Conditional random fields (CRFs) exhibits promising performance on sequential tagging tasks. It directly models the conditional probability of the label sequence without assumption on the dependencies among the observations.

Given the input sequence $\mathbf{x}=\left \{ x_{1},x_{2},...,x_{L} \right \}$ and the label sequence $\mathbf{y}=\left \{ y_{1},y_{2},...,y_{L} \right \}$, linear-chain CRFs define the conditional probability $p(\mathbf{y}|\mathbf{x})$ as
\begin{equation}
p(\mathbf{y}|\mathbf{x}) = \frac{1}{Z(\mathbf{x})}\exp(\sum_{d=1}^{D}\bm{\theta}_{d}F_{d}(\mathbf{x},\mathbf{y})),
\end{equation}
where $F_{d}(\mathbf{x},\mathbf{y})=\sum_{i=1}^{L}f_{d}(y_{i},y_{i-1},\mathbf{x},i)$ are feature functions. $Z(\mathbf{x})=\sum\nolimits_{y_{i}}\exp(\sum_{d=1}^{D}\bm{\theta}_{d}f_{d}(y_{i},y_{i-1},\mathbf{x},i))$ is the partition function and $\bm{\theta}$ is a weight vector.

Given the training dataset $X = \left \{ \mathbf{x}_{i} \right \}_{i=1}^{N}$ and $Y = \left \{ \mathbf{y}_{i} \right \}_{i=1}^{N}$ , the parameter $\bm{\theta}$  can be estimated by maximizing $\sum_{i=1}^{N}\ln p(\mathbf{y}_{i}|\mathbf{x}_{i})$ in which optimization algorithms like conjugate gradient and L-BFGS \cite{re25} can be used.

\section{Modeling sequential annotation for sequence labeling with crowds (SA-SLC)}
In this section, we describe the formulation of SA-SLC and the proposed valid label sequence inference for efficient marginalization of SA-SLC. Also, Expectation Maximization for parameter estimation is presented.

\subsection {Formulation} 
Given $N$ sequential dataset $X=\left \{ \mathbf{x}_{i} \right \}_{i=1}^{N} $ where $\mathbf{x}_{i} = \left \{ x_{j}\right \}_{j=1}^{L_{i}}$, $Y = \left \{ \mathbf{y}_{i}^{1} ,\mathbf{y}_{i}^{2},...,\mathbf{y}_{i}^{K}\right \}_{i=1}^{N}$ is the crowd label sequences tagged by $K$ different annotators. The joint conditional probability $p(Y|X)$ is defined as
\begin{equation}
\begin{aligned}
p(Y|X) &=\prod_{i=1}^{N}\prod_{k=1}^{K}p(\mathbf{y}_{i}^{k}|\mathbf{x}_i) \\
          & = \prod_{i=1}^{N}\prod_{k=1}^{K}\sum \nolimits_{\mathbf{z}_i}p(\mathbf{y}_{i}^{k},\mathbf{z}_i|\mathbf{x}_i) \\
          &= \prod_{i=1}^{N}\prod_{k=1}^{K}\sum \nolimits_{\mathbf{z}_i}p(\mathbf{y}_{i}^{k}|\mathbf{z}_i,\mathbf{x}_i)p(\mathbf{z}_i|\mathbf{x}_i),\\
\end{aligned}
\end{equation}
where $\mathbf{z}_i$ denotes one possible label sequence for the $i_{th}$ input sequence. $p(\mathbf{y}_{i}^{k}|\mathbf{z}_i,\mathbf{x}_i)$ is defined as
\begin{equation}
p(\mathbf{y}_{i}^{k}|\mathbf{z}_i,\mathbf{x}_i)=\prod_{j=1}^{L_i}p(y_{ij}^{k}|z_{ij},x_{ij}),
\end{equation}
where $L_i$ denotes the length of $i_{th}$ sequence.

Different from labeling the instances that are assumed to be independent from each other, internal label dependency among the tokens in the sequence should be considered for sequential annotation \cite{re24}. When inferring the output label of a given token, a good annotator will utilize local and non-local label dependency and then provide the most reasonable label, where $p(y_{ij}|z_{ij},x_{ij})\neq p(y_{ij}|z_{ij})$. In this paper, we formulate $p(y_{ij}|z_{ij},x_{ij})$ with parametric modeling to measure the expert level of annotators in capturing local and non-local label dependency.

First, to model the annotator's expertise in capturing local label dependency, $p(y_{ij}|z_{ij},x_{ij})$ is defined as
\begin{equation}
 p(y_{ij}|z_{ij},x_{ij})=p(y_{ij}|,y_{i(j-1)},z_{ij}),
\end{equation}
where $y_{ij}$ depends on $z_{ij}$ and the annotator's previous annotation $y_{i(j-1)}$. For example, in a task of named entity recognition, as shown in Figure 2, a good annotator will not assign ``I" to 
the word ``carrier" if the previous annotation for ``national" is ``O". When the word ``Bosnian" is tagged with ``B-ORG", ``Association" is more likely to be annotated with ``I-ORG".

For the $k_{th}$ annotator, $y_{ij}^{k}$ is assumed to follow categorical distribution given $(y_{i(j-1)},z_{ij})$, which is expressed as
\begin{equation}
y_{ij}^{k}\sim Cat(\bm{\alpha} _{1}^{k},\bm{\alpha }_{2}^{k},...,\bm{\alpha} _{M}^{k}),
\end{equation}
where $M$ is the number of all possible labels. The expert level of capturing local dependency of $k_{th}$ annotator is modeled by a $M\times M\times M$ confusion matrix $\bm{\alpha}^k$. $\bm{\alpha}^k(i,j,h)$ is defined as
\begin{equation}
\bm{\alpha}^k(i,j,h)=p(y_l=h|z_l=j,y_{l-1}=i),
\end{equation}
where the ground-truth is $j$, previous annotation is $i$ and the assigned label is $h$.

\begin{figure*}
\centering
\includegraphics[width=5.0in,height = 2.5 in ]{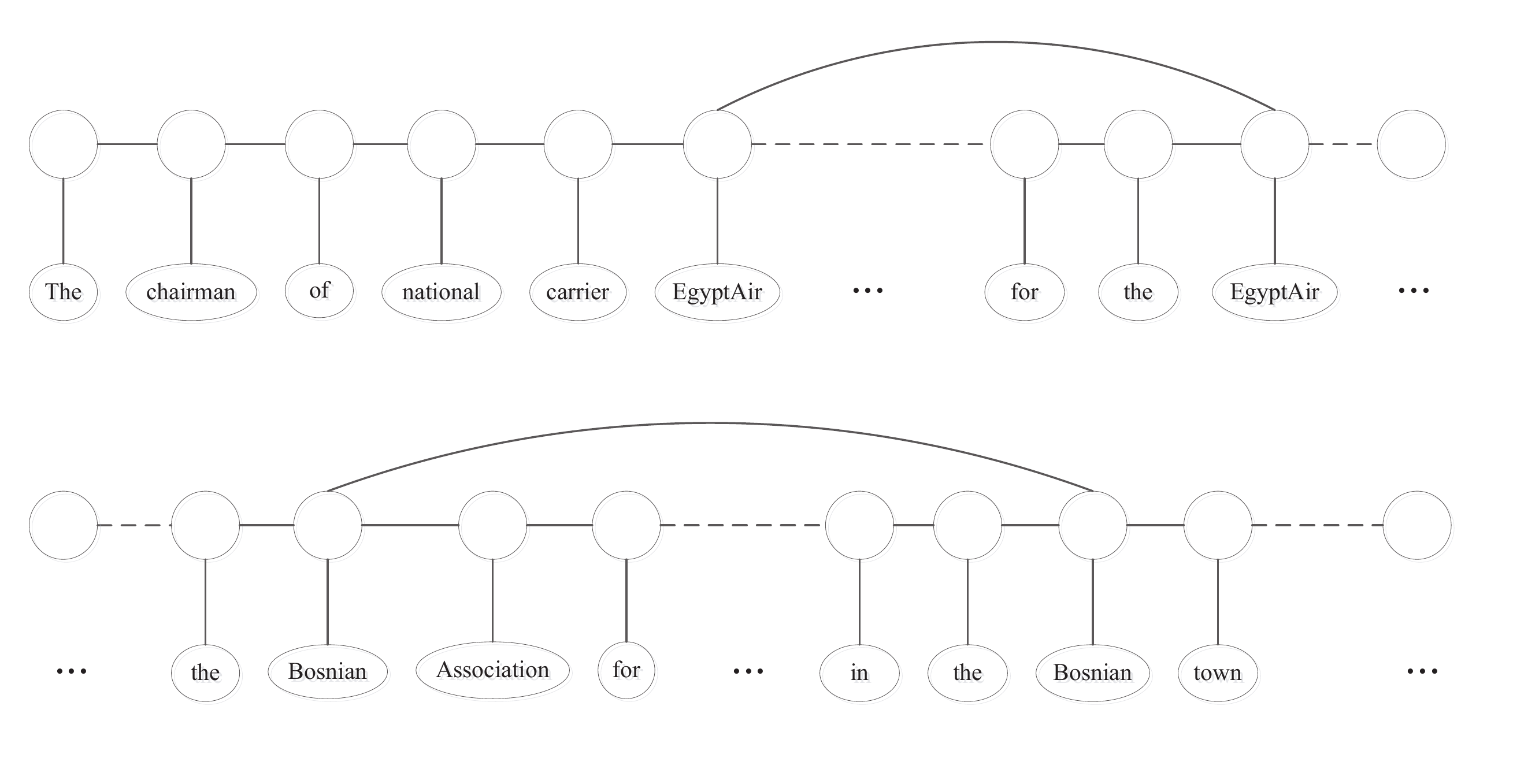}
\caption {Two examples of non-local label dependency.}

\end{figure*}

In sequence labeling, encoding non-local label dependency further improves label consistency of the whole sequence \cite{re9, re26}. For sequential annotation, the annotator naturally refers to the same (or similar) mention in previous annotation when deciding the label of  the current token. As shown in Figure 2, the annotation for the word ``EgyptAir" (or ``Bosnian") can be a good indicator for labeling the subsequent occurrence ``EgyptAir" (or ``Bosnian"). In this paper, we focus on modeling the non-local label dependency between the same mentions, which is expressed as 
\begin{equation}
\begin{aligned}
p(y_{ij}|z_{ij},x_{ij}) &= p(y_{ij}|z_{ij}, [x_{ij}, (x',y')])\\
                                    &= p(y_{ij}|y_{ij}^{'},z_{ij}),
\end{aligned}
\end{equation}
where $x'$ is the same mention of $x_{ij}$ in the previous sequence. Then $x_{ij}$ has the pseudo label $y_{ij}^{'}$ which equals the label $y'$ assigned to $x'$. Since $x_{ij}$ may have several same mentions, we choose the one whose position is closest to $x_{ij}$.

Similarly, $y_{ij}^{k}$ follows categorical distribution given $(y_{ij}^{'},z_{ij})$, then the expert level of capturing non-local label dependency of $k_{th}$ annotator is modeled by a $M\times M\times M$ confusion matrix $\bm{\beta}^k$. $\bm{\beta}^k(i,j,h)$ is defined as
\begin{equation}
\bm{\beta}^k(i,j,h)=p(y_l=h|z_l=j,y_{l}^{'}=i),
\end{equation}
where the ground-truth is $j$, previous same mention is $i$ and the assigned label is $h$.

Based on the above description, $p(\mathbf{y}_{i}^{k}|\mathbf{z}_i,\mathbf{x}_i)$ is defined as
\begin{equation}
\begin{aligned}
&p(\mathbf{y}_{i}^{k}|\mathbf{z}_i,\mathbf{x}_i)\\
&=\prod_{j=1}^{L_i}p(y_{ij}^{k}|z_{ij},x_{ij})\\
&=\prod_{j=1}^{L_i}\scalebox{0.85}{$p(y_{ij}|y_{ij}^{'},z_{ij})^{\llbracket \exists x_{ij}\in\mathbf{ x}_{i}{(0,j-1)}\rrbracket  }p(y_{ij}|,y_{i(j-1)},z_{ij})^{1-\llbracket \exists x_{ij}\in\mathbf{ x}_{i}{(0,j-1)}\rrbracket } $} ,
\end{aligned}
\end{equation}
where $\exists x_{ij}\in\mathbf{ x}_{i}{(0,j-1)}$ denotes that there exists at least one $x_{ij}$ in the sequence $\mathbf{ x}_{i}{(0,j-1)}$. $\llbracket P \rrbracket  = 1$ if $P$ is true and zero otherwise.

To improve efficiency of parameter estimation and provide a reasonable modeling of label dependency, in Equation (11) the proposed SA-SLC ignores local label dependency when there is a previous occurrence of $x_{ij}$. Utilizing local and non-local label dependency simultaneously requires a four-dimensional confusion matrix with size $M\times M\times M\times M$ that encodes label information from previous same mention, previous annotation, the ground-truth and the assigned label, which greatly increases the size of parameters to be estimated.

\subsection{Decoding: Valid label sequence inference}
The joint conditional probability, as shown in Equation (4), marginalizes all possible ground-truth label sequences out. But finding the joint conditional probability may not be practical because it is computationally expensive especially the number of these sequences is exponential to the length of sequence. For example, in the task of named entity recognition, for a sentence with 10 words, the total number of possible label sequences can be $9^{10}$. Besides, for sequence annotation, enumerating all possible label sequences will generate large number of invalid sequences, for example, possible label sequences for the sequence in Figure 3 could be ``O O O I-ORG O I-ORG I-ORG" and ``O O O B-PER I-ORG O I-ORG", which makes no sense for sequence analysis.

\begin{figure}[H]
\centering
\includegraphics[width=3in,height = 1.2in ]{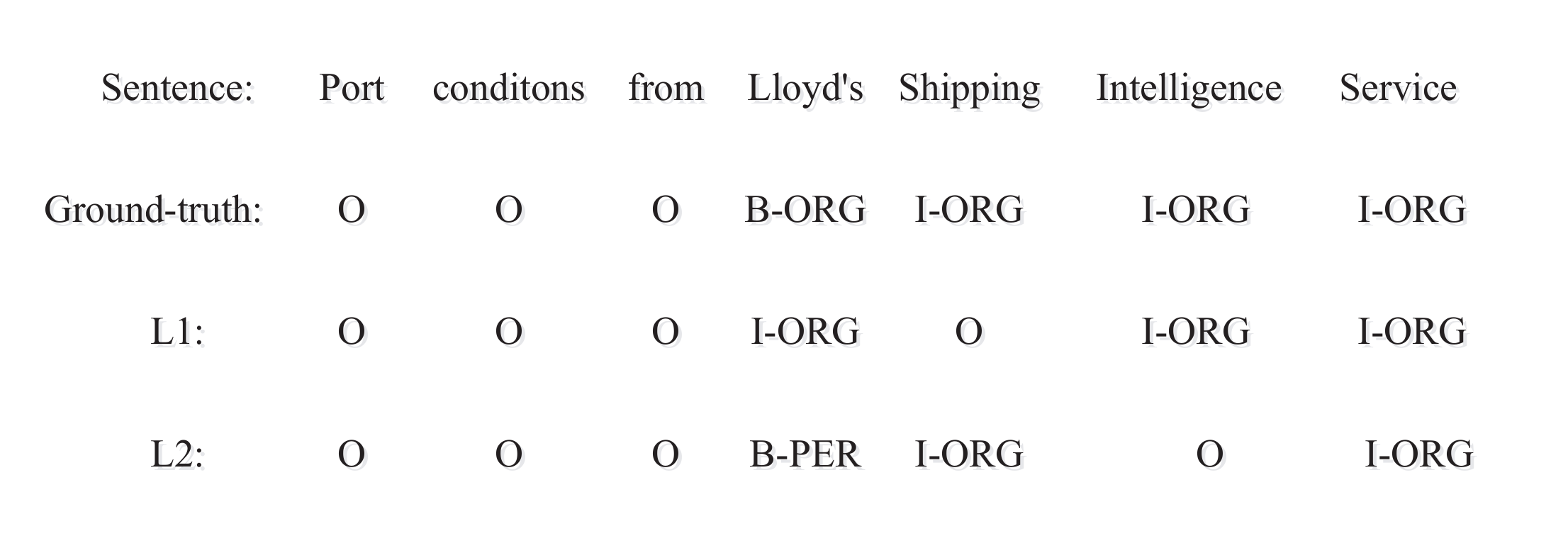}
\caption {Example of possible ground-truth label sequences for ``Port conditions from Lloyd's Shipping Intelligence Service".}
\label{fig:secondfigure}
\end{figure}

In this section, we describe a newly developed method called valid label sequences inference (VLSE) to derive the valid ground-truth label sequences from crowd sequential annotations. It estimates the possible ground-truth labels in the token-wise level by exploring the label consistency among annotators. VLSE, then decodes the valid label sequences for the whole input sequence.

First, let $\left \{ l_r\right \}_{r=1}^R$ denotes the set of unique labels collected from $K$ annotations where $R$ is the total number of unique labels. We define the degree of label consistency  $LC_{ij}$  for the $j_{th}$ element in the $i_{th}$ sequence as
\begin{equation}
LC_{ij}=\frac{\max\limits_{l_r}\left ( \left \{ n_{l_r} \right \}_{r=1}^{R} \right )}{R},
\end{equation}
where $n_{l_r}$ represents occurrence times of the label $l_r$ in K annotations. 

To derive the possible ground-truth labels $L_{ij}$ for the $j_{th}$ element in the $i_{th}$ sequence, we consider three different cases which is expressed as
\begin{equation}
L_{ij} = 
\left\{
\begin{array}{lcl}
\left \{  \arg \max\limits_{l_r}\left ( \left \{ n_{l_r} \right \}_{r=1}^{R} \right )\right \}, if  \; LC_{ij} \geq T_1 
 ,\\
\left \{ l_{r} \right \}_{r=1}^{R}, if  \; T_2 < LC_{ij} <  T_1,\\
\left \{ l_{m} \right \}_{m=1}^{M}, if \; LC_{ij} \leq  T_2,
\end{array} \right.
\end{equation}
where $M$ is the number of all possible labels. By specifying appropriate threshold $T_1$ and $T_2$, three cases of obtaining the set of possible true labels $L_{ij}$ are as follows: If $LC_{ij}$ is high, the label with majority vote is chosen; if $LC_{ij}$ is moderate, the set of all labels assigned by the annotators to $x_{ij}$ is considered; if $LC_{ij}$ is low, all available labels in the tagging scheme are considered.


Generally, differences remain among K annotations. Annotators are more likely to reach an agreement when tagging simple data, where $LC_{ij} $ can be K. For the element in the sequence with much lower $LC$, all possible labels will be considered as the ground-truth labels. For example, as shown in Figure 4, there are five annotators tagging the sequence ``--Dimitris Kontogiannis, Athens Newsroom +301 3311812-4". The possible ground-truth labels for each token can be derived based on Equation (13), which is presented in Figure 5.

\begin{figure}[H]
\centering
\includegraphics[width=3.5in,height = 1.5in ]{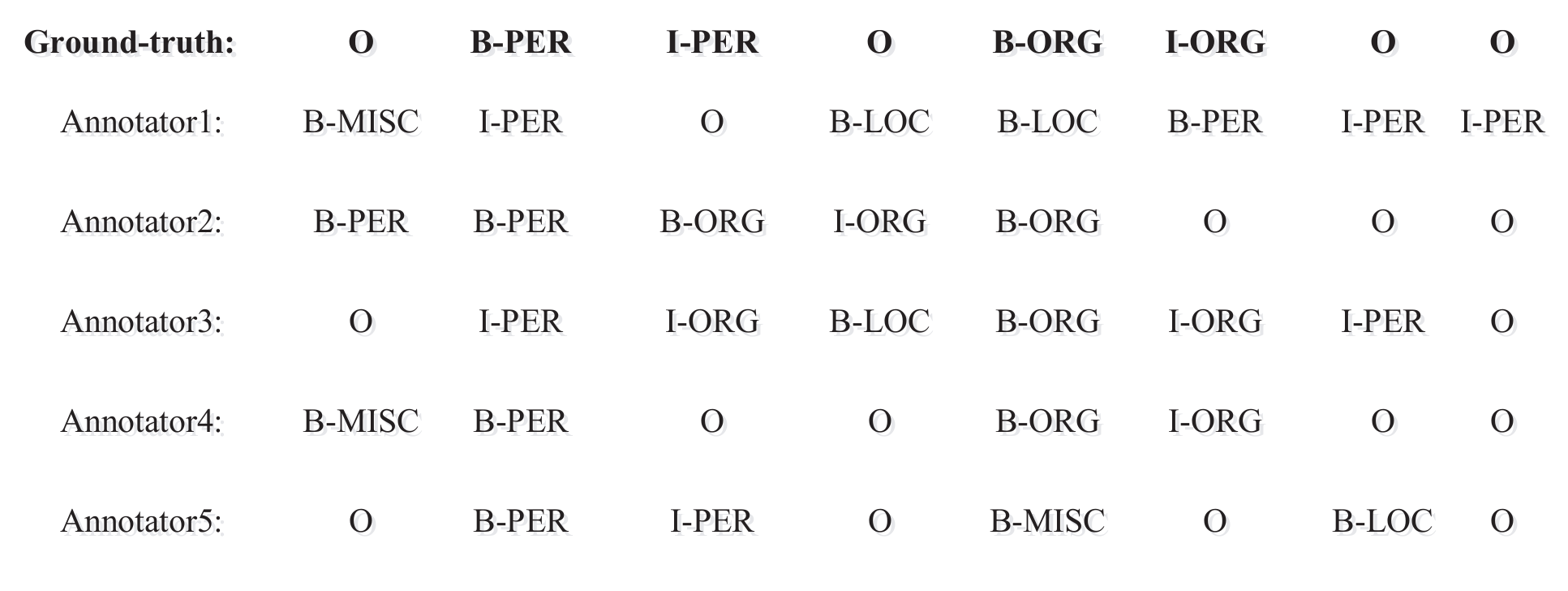}
\caption {An example with five annotators. }
\label{fig:secondfigure}
\end{figure}

\begin{figure}[H]
\centering
\includegraphics[width=3.5in,height = 1.5in ]{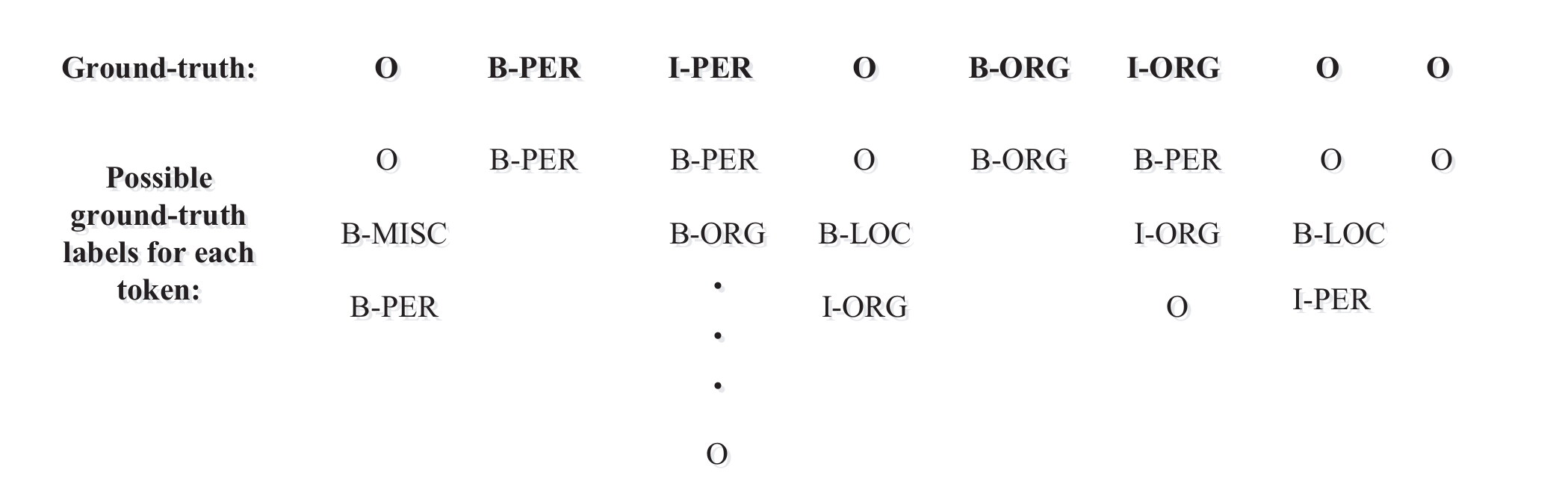}
\caption {An example: the possible ground-truth labels for each token derived by VLSE. }
\label{fig:secondfigure}
\end{figure}

To derive the valid label sequences, based on the determined possible ground-truth labels for each element,  VLSE further prunes some sub-paths that violates the constraints created from prior knowledge. For example, for a task of named entity with ``PERSON, ORGANIZATION and LOCATION", the constraints can be { ``I-" cannot follow ``O", next to ``B-ORG" could be ``B-" , ``O" and ``I-ORG" (same to all ``B-") , ``I-PER" only follow ``B-PER" or ``I-PER" (same to ``I-") }. Let $P_{t}(s_{i})$ denote the possible sub-paths from the previous token $t-1$ to the state $s_i$ (i.e. label) in the current token $t$, which is defined as
\begin{equation} 
P_{t}(s_{i})=\left \{ P_{t-1}(s_{j})\right \}_{j=1}^{T},
\end{equation}
where $\left \{ s_{j} \right \}_{j=1}^{T}$ is the set of possible states in the token $t-1$, and the transition $s_{j}\rightarrow s_{i}$ does not violate the constraints. 

By incorporating the constraints into the forward inference for label sequence derivation, VLSE further reduces the number of candidate label sequences in the marginalization which results in improving the quality of possible ground-truth label sequences. For example, as shown in Figure 6, without pruning the number of possible ground-truth label sequences is 729. Based on the constraints described above for the task of named entity recognition with ``PERSON, ORGANIZATION and LOCATION", the sub-paths with red dotted line will be pruned. As a result, some invalid label sequences, such as ``O B-PER I-LOC I-ORG B-ORG O I-PER O"  and ``O B-PER I-ORG I-ORG B-ORG I-ORG I-PER O", can be excluded. It is worth noting that the number of possible ground-truth label sequences is decreased to 315, which is about half the size of original number without pruning. This pruning process will be highly significant in obtaining the valid possible ground-truth label sequences.

\begin{figure}[H]
\centering
\includegraphics[width=3.5in,height = 2in ]{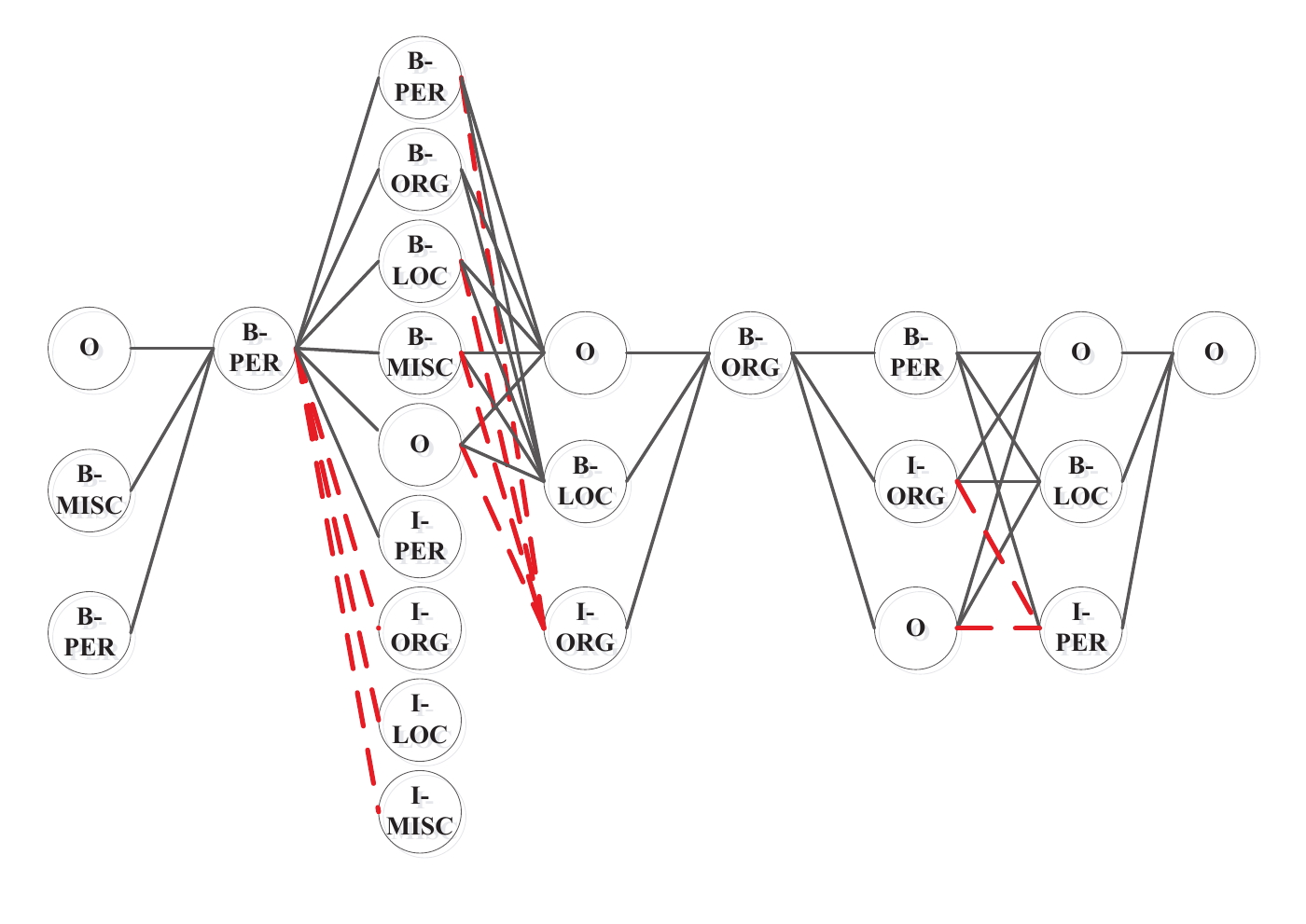}
\caption {An example: prune subpaths that violate the constraints by VLSE.}
\label{fig:secondfigure}
\end{figure}

\subsection{Parameter Estimation}
The objective function is defined as
\begin{equation}
\log p(Y|X; \bm{\theta}) =\sum_{i=1}^{N}\sum_{k=1}^{K}\log \sum \nolimits_{\mathbf{z}_i}p(\mathbf{y}_{i}^{k}|\mathbf{z}_i,\mathbf{x}_i)p(\mathbf{z}_i|\mathbf{x}_i; \bm{\theta}),
\end{equation}
where $\bm{\theta}$ represents the weights vector for feature function.

The set of parameters $\left \{ \bm{\alpha },\bm{\beta}, \bm{\theta} \right \}$ can be estimated by maximizing the above likelihood function. Since the ground-truth label sequence is missing, we employ the Expectation Maximization (EM) \cite{re27} to solve likelihood maximization, which is detailed as follows:

\textbf{Initialization}: Let $\bm{\alpha}_{ij}^k$ denotes the $j_{th}$ row of $i_{th}$ layer of $\bm{\alpha}^k$, we sample $\bm{\alpha}_{ij}^k$ from a common Dirichlet prior with $\bm{\alpha}_{ij}^k \sim Dir(\mathbf{1})$. Similarly, $\bm{\beta}_{ij}^k$ is sampled from the Dirichlet distribution with $\bm{\beta}_{ij}^k\sim Dir(\mathbf{1})$. Further, we random select an annotator and use the corresponding labelings as pseudo ground-truths to initialize $\bm{\theta}$.

\textbf{E-step}: compute $\widetilde{p}(\mathbf{z}_{i})$ for each valid ground-truth label sequence $\mathbf{z}_{i}$.
\begin{equation}
\begin{aligned}
\widetilde{p}(\mathbf{z}_{i}) &=p(\mathbf{z}_{i}|\mathbf{x}_{i},\mathbf{y}_{i}) \\
&= \frac{\prod_{k=1}^{K} p(\mathbf{y}_{i}^{k}|\mathbf{z}_i,\mathbf{x}_i)p(\mathbf{z}_i|\mathbf{x}_i)}{\sum \nolimits_{\mathbf{z}_i}\prod_{k=1}^{K}p(\mathbf{y}_{i}^{k}|\mathbf{z}_i,\mathbf{x}_i)p(\mathbf{z}_i|\mathbf{x}_i; \mathbf{\theta})}.
\end{aligned}
\end{equation}

\textbf{M-step}: maximize $\sum_{i=1}^{N}\sum_{k=1}^{K}E_{\widetilde{p}(\mathbf{z}_{i})}\left [ \log p(\mathbf{y}_{i}^{k},\mathbf{z}_{i}|\mathbf{x}_{i}) \right ]$.

The expectation over $\log p(\mathbf{y}_{i}^{k},\mathbf{z}_{i}|\mathbf{x}_{i})$ can be transformed into
\begin{equation}
\begin{aligned}
&\sum_{i=1}^{N}\sum_{k=1}^{K}E_{\widetilde{p}(\mathbf{z}_{i})}\left [ \log p(\mathbf{y}_{i}^{k},\mathbf{z}_{i}|\mathbf{x}_{i}) \right ]
=\sum_{i=1}^{N}E_{\widetilde{p}(\mathbf{z}_{i})}\left [ \log p(\mathbf{z}_i|\mathbf{x}_i) \right ]  \\ 
& + \sum_{i=1}^{N}\sum_{k=1}^{K}E_{\widetilde{p}(\mathbf{z}_{i})}\left [ \log p(\mathbf{y}_i^{k} | \mathbf{x}_i,\mathbf{z}_i)\right ].
\end{aligned}
\end{equation}

The parameter $\bm{\theta}$ is obtained by maximizing $\sum_{i=1}^{N}E_{\widetilde{p}(\mathbf{z}_{i})}\left [ \log p(\mathbf{z}_i|\mathbf{x}_i) \right ]$, which is expressed as
\begin{equation}
\begin{aligned}
&\sum_{i=1}^{N}E_{\widetilde{p}(\mathbf{z}_{i})}\left [ \log p(\mathbf{z}_i|\mathbf{x}_i) \right ]   =\sum_{i=1}^{N}\sum \nolimits_{\mathbf{z}_i}\widetilde{p}(\mathbf{z}_{i})\log p(\mathbf{z}_i|\mathbf{x}_i) \\
& = \sum_{i=1}^{N}\sum \nolimits_{\mathbf{z}_i}\widetilde{p}(\mathbf{z}_{i}) \left [ \sum_{d=1}^{D}\theta_{d}F_{d}(\mathbf{x}_i,\mathbf{z}_i)-\log(Z(\mathbf{x}_i)) \right ],
\end{aligned}
\end{equation}
where the parameter optimization is similar to the basic CRF model that is described in Section \Romannum{3} B.

Then maximum likelihood parameter estimation for categorical distribution is employed to obtain the parameters $\left \{ \bm{\alpha} , \bm{\beta } \right \}$.

\section{Experiments}
In this section, we evaluate the performance of SA-SLC on the NLP datasets related with news and biomedicine by comparison with baselines and parameters' sensitivity analysis (i.e. the number of annotators and annotators' level of expertise). Apart from the four NLP datasets containing both gold and simulated crowdsourced label sequences, we also apply SA-SLC to the crowd dataset collected from Amazon’s Mechanical Turk to investigate its effectiveness in real-world application.

\subsection{Datasets}

CoNLL 2003 \cite{re28}: The dataset, consisting of over 21,000 sentences, is used for named entity recognition, which includes four types of named entities: persons, organizations, locations and names of miscellaneous. 

BC2GM \cite{re29}: This dataset consists of 20,000 sentences collected from biomedical publication abstracts and the annotations have two entity types (i.e. Gene or Protein).

NCBI-disease \cite{re30}: The NCBI Disease corpus consists of 793 PubMed abstracts and is used to extract disease mentions. The abstract is fully annotated at the mention of diseases.

JNLPBA \cite{re31}: The Joint workshop on NLP in Biomedicine and its Applications corpus consists of 2,404 publication abstracts, which is annotated for five entity types: DNA, RNA, cell line, cell type and protein.

The above datasets only contain gold label sequences. To simulate crowd labels from multiple annotators, we choose precision to present annotators' reliability as accuracy is not the best measure for imbalanced datasets. Then crowd labels are generated by controlling the average precision $p$ among $K$ annotators, where a correct named entity should exactly match the gold entity. The details of four datasets are summarized in Table \uppercase\expandafter{\romannumeral1}.

\begin{table}[h]
\centering
\caption{Summary of the datasets}
\begin{tabular}{lcccc} 
\hline
\multicolumn{1}{c}{\multirow{2}{*}{Dataset}} & \multicolumn{3}{c}{\#sentences or abstracts} & \multirow{2}{*}{\#labels}  \\
\multicolumn{1}{c}{}                         & train & dev  & test                          &                            \\ 
\hline
CoNLL 2003                                   & 6000  & 2000 & 3453                          & 9                          \\
BC2GM                                        & 8000  & 4000 & 5038                          & 3                          \\
NCBI-disease                                 & 400   & 150  & 200                           & 3                          \\
JNLPBA                                       & 1000  & 500  & 500                           & 11                         \\
\hline
\end{tabular}
\end{table}

\subsection{Baselines}
We select multiple baselines that can be grouped into two types: wrapper methods and joint models. Wrapper methods directly infer the ground-truths from crowds and then input these labels to a sequence labeling model, while joint models estimate annotator’s expertise and the classifier simultaneously \cite{re32}. The baselines are described as follows:

MVtoken \cite{re20}: The ground-truth label sequence is obtained by choosing the label with more votes at the token level.

DS (David-Skene) \cite{re6}: The EM algorithm is employed to assign weight to each vote at the token level.

MACE \cite{re33}: By including a binary latent variable that denotes if and when each annotator is spamming, the model can identify which annotators are trustworthy and produce the true label.

Sembler \cite{re24}: The model focuses on crowd sequential labeling, which is the extension of crowdsourcing problem on the instances that are independent from each other.

HMM-Crowd \cite{re22}: Based on HMMs, HMM-Crowd further models the ``crowd component" by including additional parameters for measuring the reliability of annotators.

In the experiments, MVtoken, DS and MACE are in the category of wrapper methods. To be fair, we choose CRFs as the subsequent sequence labeling model for these methods.

\subsection{Experimental results}
In this section, we follow the evaluation measure employed by CoNLL 2003 with
\begin{equation}
F1=\frac{2\times precision \times recall}{precision+recall},
\end{equation}
where precision is the percentage of named entities detected by the learning model while recall is the percentage of named entities present in the dataset that are detected by the learning model, and only a named entity that exactly matches the gold can be counted as the correct one.

First, we choose the number of annotators and annotating precision in medium level to approximate the real setting of crowd annotations in Amazon’s Mechanical Turk (AMT) \cite{re20}, where $K=5$ and $p=0.5$. Table \uppercase\expandafter{\romannumeral2} and \uppercase\expandafter{\romannumeral3} report the mean and standard deviation of F1 score with 10-fold cross-validation on the combination of training and development datasets. Further, pairwise t-test at 5\% significance level is recorded. We can observe that the proposed SA-SLC achieves the best performance in BC2GM, NCBI-disease and JNLPBA datasets, and it significantly outperforms MACE, MVtoken and Sembler in all selected datasets.

\begin{table}[H]
\centering
\caption{F1 score (mean$\pm$std) of each comparing algorithms on CoNLL 2003 and BC2GM datasets (\%).}
\begin{threeparttable}
\begin{tabular}{lcc} 
\hline
Model     & CoNLL 2003 & BC2GM  \\ 
\hline
MVtoken   & $63.09\pm4.70$ $\bullet $          &   $59.06\pm0.58$  $\bullet $    \\
DS        &   $70.02\pm 4.35$  $\circ $        &     $58.33\pm1.17$ $\bullet $   \\
MACE      &   $64.21\pm6.67$ $\bullet $         &     $61.98\pm0.71$ $\bullet $   \\
Sembler   &   $63.71\pm4.95$  $\bullet $        &     $63.59\pm1.16$  $\bullet $  \\
HMM-Crowd & $69.17\pm4.39$  $\circ $         &  $61.65\pm1.25$ $\bullet $      \\
SA-SLC    &   $69.74\pm5.13$          &     $64.71\pm1.07$   \\
\hline
\end{tabular}
\begin{tablenotes}
 \item[] $\bullet $/$ \circ  $ denotes whether the performance of SA-SLC is statistically superior/inferior to the comparing methods (pairwise t-test at 5\% significance level).
 \end{tablenotes}
 \end{threeparttable}
\end{table}

\begin{table}[H]
\centering
\caption{F1 score (mean$\pm$std) of each comparing algorithms on NCBI-disease and JNLBPA datasets (\%).}
\begin{threeparttable}
\begin{tabular}{lcc} 
\hline
Model     & NCBI-disease & JNLPBA  \\ 
\hline
MVtoken   &  $73.75\pm2.81$  $\bullet $         &   $53.76\pm1.84$ $ \bullet  $     \\
DS        &  $72.79\pm3.19$  $\bullet $         &   $51.54\pm0.95$ $\bullet $     \\
MACE      &  $72.41\pm2.55$ $\bullet $          &   $56.26\pm1.96$ $\bullet $     \\
Sembler   &   $74.79\pm3.25$ $\bullet $         &   $54.27\pm2.06$ $ \bullet $     \\
HMM-Crowd &   $74.08\pm3.16$ $ \bullet  $         &  $54.18\pm 2.06$ $\bullet $      \\
SA-SLC    &   $75.59\pm2.78$          &    $58.14\pm0.94$     \\
\hline
\end{tabular}
\begin{tablenotes}
 \item[] $\bullet $/$ \circ  $ denotes whether the performance of SA-SLC is statistically superior/inferior to the comparing methods (pairwise t-test at 5\% significance level).
 \end{tablenotes}
 \end{threeparttable}
\end{table}

Then, to predict label sequences of test data, we repeat sampling crowd annotations three times with fixed $p$ and record the average performance. Figure 7 to 10 summarize the F1 score of each comparing method on testing set of four datasets, where $p$ varies from 0.9 to 0.1. The following observations can be made:

\begin{itemize}

\item
For CoNLL 2003 dataset, the performance of each comparing method is significantly positively correlated with $p$. When $p$ is lower than 0.5, the reliability of crowd annotations is greatly reduced, which degrades the performance of the subsequent learning and prediction. It is worth noting that MVtoken and DS outperform the other comparing models in most cases but joint models (Sembler, HMM-Crowd and SA-SLC) perform stably with the lower $p$. There is a certain degree of randomness in simulating annotators with different expertise. Medium level configuration (e.g. $p=0.5$ and $p=0.7$) may benefits the simple learning mechanism of MV and DS, which results in improving the quality of inferred ground-truths.

\item
As shown in Figure 8, SA-SLC outperforms the other comparing methods on BC2GM dataset, the joint models (i.e. SA-SLC, Sembler and HMM-Crowd) for crowdsourced learning always achieve better results than the wrapper methods (e.g. MVtoken and DS). But there is no positive correlation between $p$ and F1 score for most of comparing methods. The major reason is that annotators have not many choices as the size of label set for BC2GM is small and increasing annotators' precision doesn't significantly improve the quality of inferred ground-truths.

\item
For NCBI-disease dataset, SA-SLC always achieves the best performance. The F1 score obtained by the most of comparing methods is positively correlated with $p$. Compared with MVtoken and Sembler, SA-SLC performs stably with varying $p$. 

\item
For JNLPBA dataset, as shown in Figure 10, SA-SLC achieves comparable results with MVtoken and always outperforms the other two joint models (i.e. Sembler and HMM-Crowd). When $p$ is lower than 0.5, SA-SLC obtains the better results than the other comparing methods. Furthermore, most of comparing methods achieves higher F1 score by increasing $p$.

\end{itemize}

\begin{figure}[h]
\centering
\includegraphics[width=3.7in,height = 2.5in ]{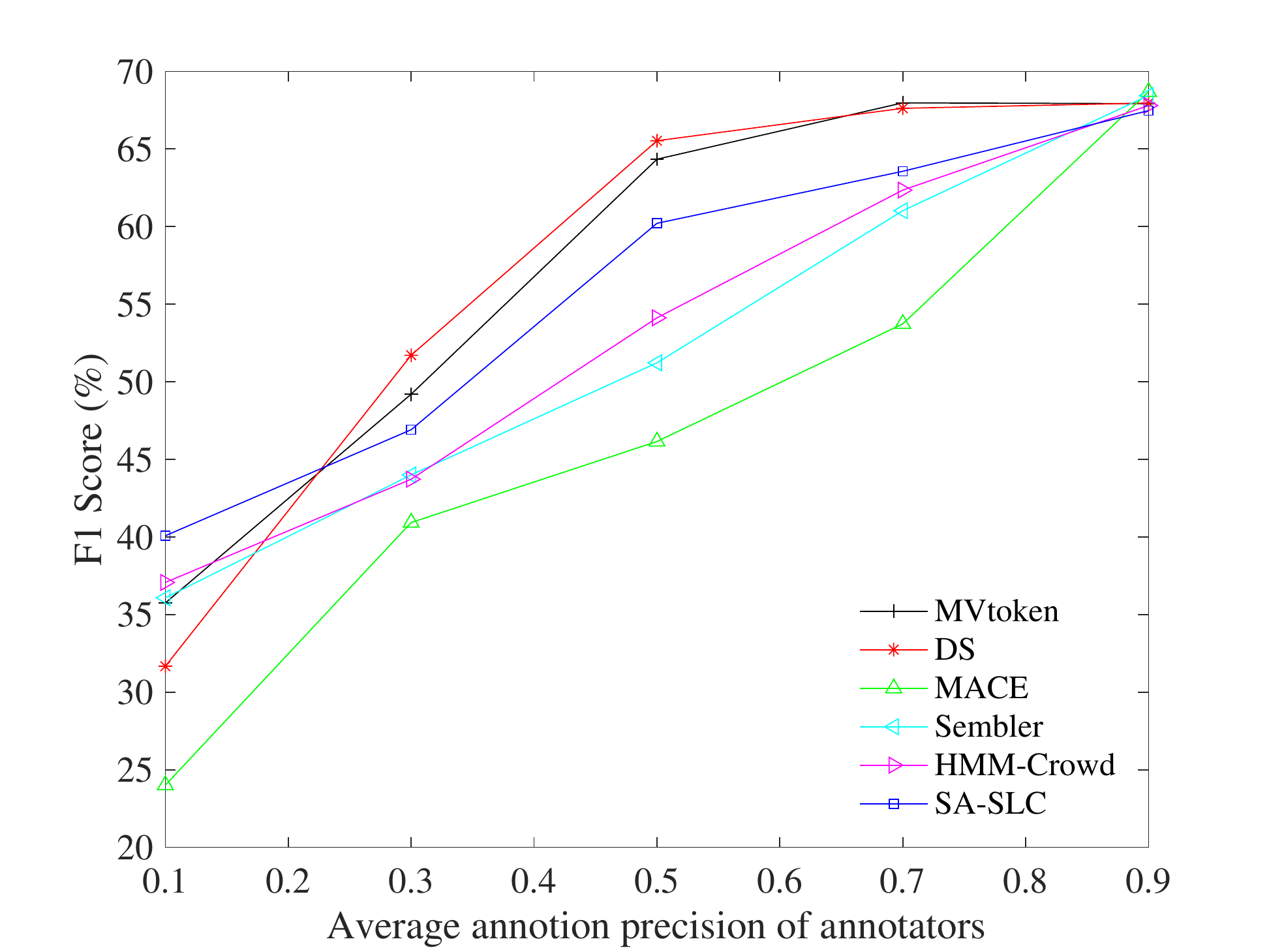}
\caption {The performance of comparing methods on CoNLL 2003 dataset.}
\label{fig:secondfigure}
\end{figure}

\begin{figure}[h]
\centering
\includegraphics[width=3.7in,height = 2.5in ]{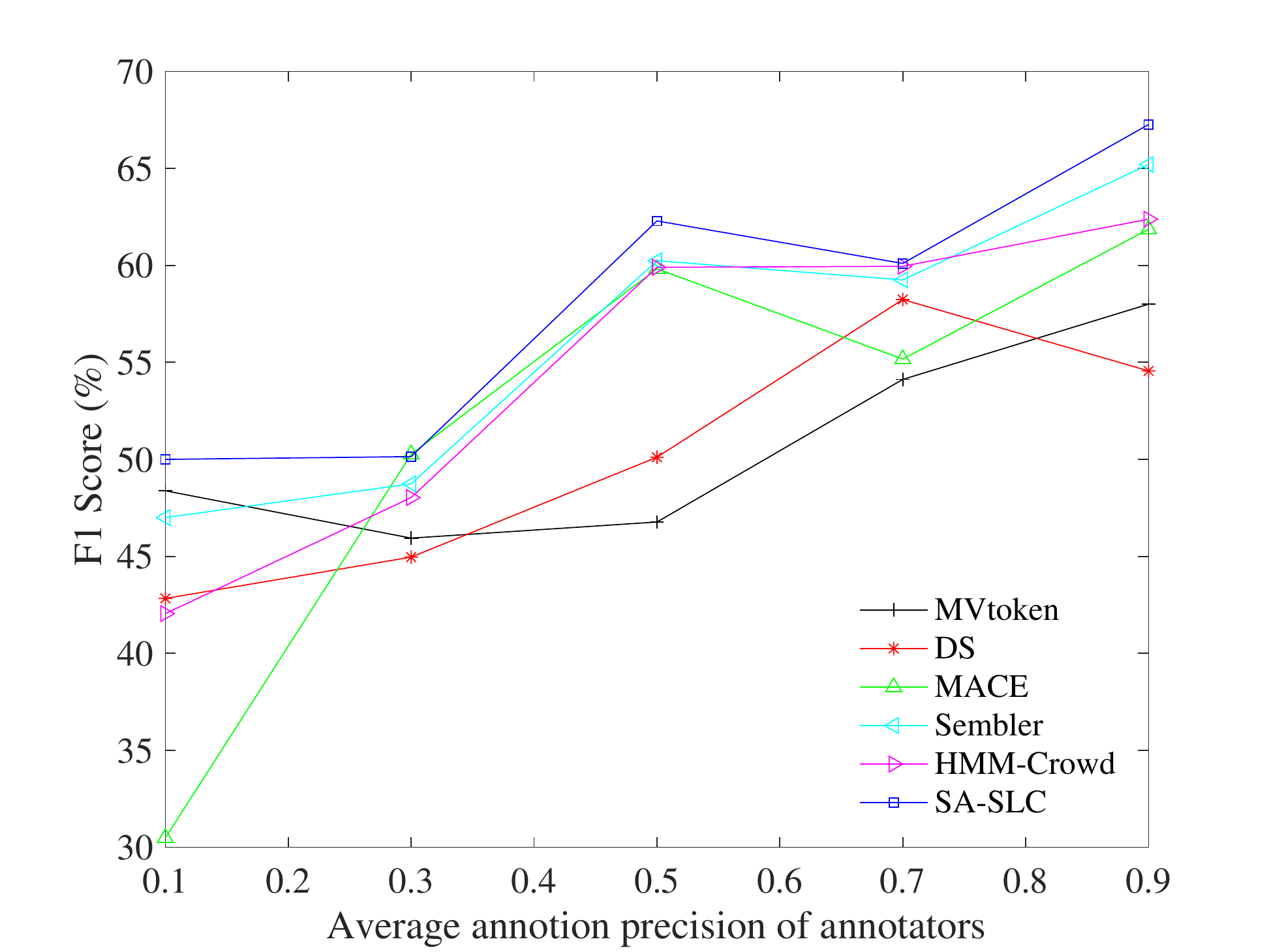}
\caption {The performance of comparing methods on BC2GM dataset.}
\label{fig:secondfigure}
\end{figure}

\begin{figure}[h]
\centering
\includegraphics[width=3.7in,height = 2.5in ]{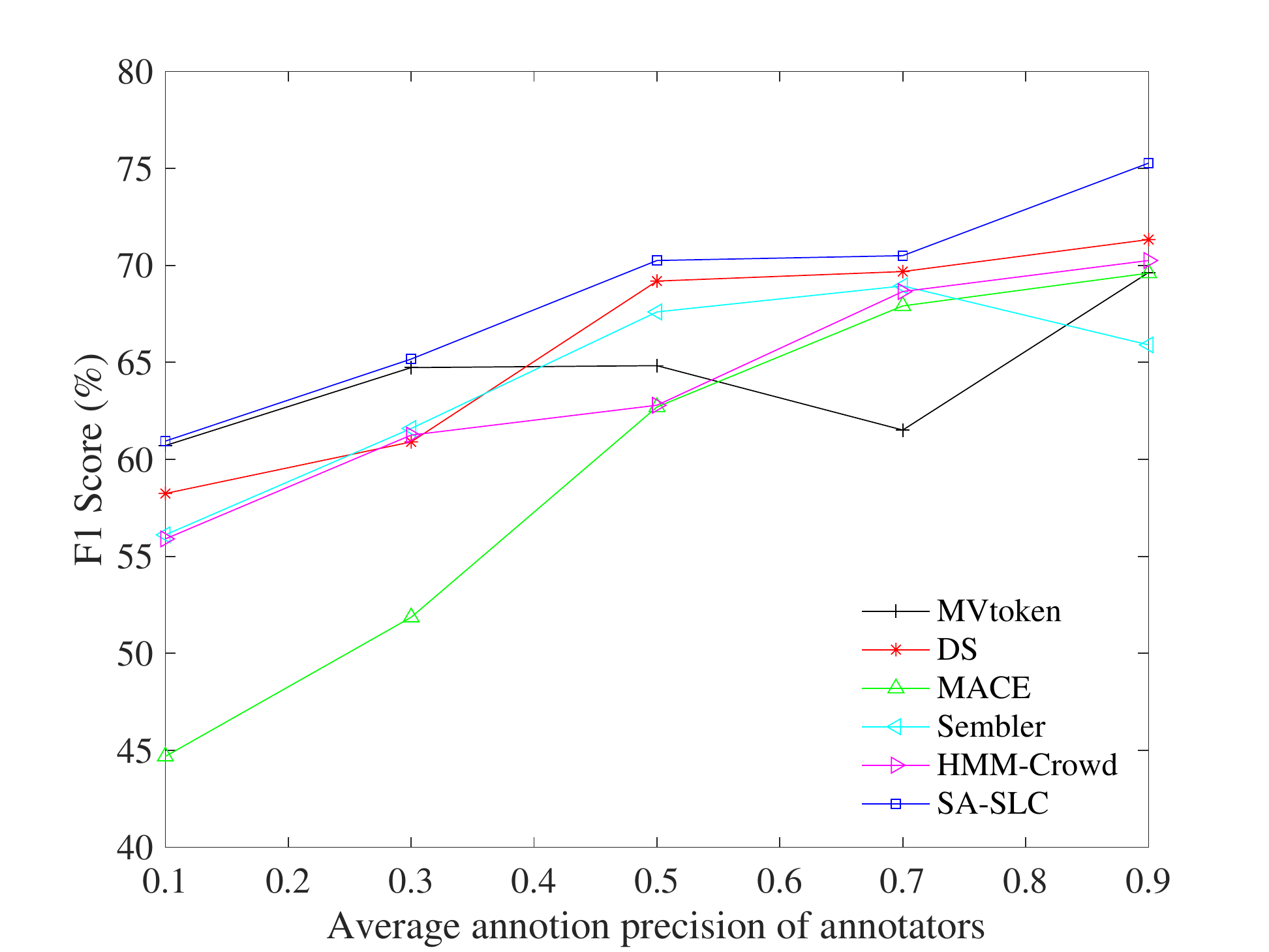}
\caption {The performance of comparing methods on NCBI-disease dataset.}
\label{fig:secondfigure}
\end{figure}

\begin{figure}[h]
\centering
\includegraphics[width=3.7in,height = 2.5in ]{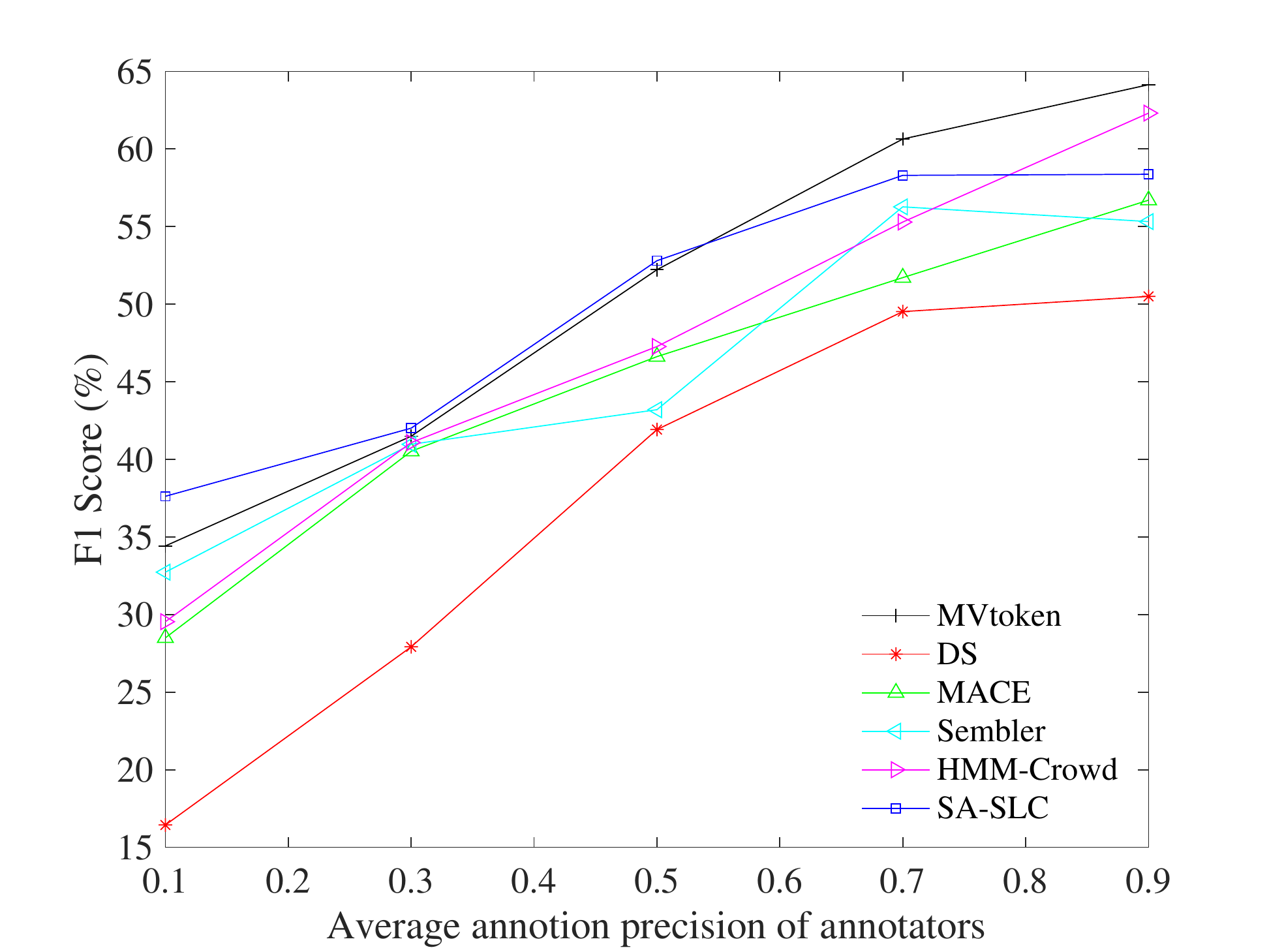}
\caption {The performance of comparing methods on JNLPBA dataset.}
\label{fig:secondfigure}
\end{figure}

Meanwhile, we observe that SA-SLC performs considerably better on the dataset with small label set (e.g. BC2GM) than that with various labels (e.g. CoNLL 2003). Figure 11 to 14 demonstrate the estimated $\bm{\alpha}$ and $\bm{\beta}$ of five annotators for the dataset CoNLL 2003 and BC2GM, where $p$ is selected as 0.5. Considering the size of parameter matrix of $\bm{\alpha}$ and $\bm{\beta}$, we present some parts of it. For example, for the dataset CoNLL 2003, as shown in Table \uppercase\expandafter{\romannumeral1}, there are 9 labels. We choose one vector which is used to describe the annotator's expertise in capturing local label dependency with the previous annotation and the ground-truth are fixed with ``B-ORG" and ``I-ORG" respectively. The selected vectors from $\bm{\alpha}$ and $\bm{\beta}$ parameter matrix for the dataset CoNLL 2003 and BC2GM are listed in Table \uppercase\expandafter{\romannumeral4}, \uppercase\expandafter{\romannumeral5}, \uppercase\expandafter{\romannumeral6} and \uppercase\expandafter{\romannumeral7}, where each entry represents annotator's expertise in capturing internal dependency.

\begin{table*}
\centering
\caption{CoNLL2003: the selected vector from  $\bm{\alpha}$ parameter matrix }
\begin{tabular}{|c|c|c|c|c|c|c|c|c|c|} 
\hline
\multirow{2}{*}{\begin{tabular}[c]{@{}c@{}}Ground-truth = ``I-ORG"\\Previous annotation = ``B-ORG"\\ \end{tabular}} & ``B-ORG" & ``I-ORG" & ``B-PER"& ``I-PER" & ``B-LOC"& ``I-LOC"& ``B-MISC"& ``I-MISC" & ``O"  \\ 
\cline{2-10}
                                                                                                                                & $\bm{\alpha}_1$     & $\bm{\alpha}_2$     & $\bm{\alpha}_3$     & $\bm{\alpha}_4$    & $\bm{\alpha}_5$     & $\bm{\alpha}_6$     & $\bm{\alpha}_7$      & $\bm{\alpha}_8$      & $\bm{\alpha}_9$  \\
\hline
\end{tabular}
\end{table*}

\begin{table*}
\centering
\caption{CoNLL2003: the selected vector from $\bm{\beta}$ parameter matrix }
\begin{tabular}{|c|c|c|c|c|c|c|c|c|c|} 
\hline
\multirow{2}{*}{\begin{tabular}[c]{@{}c@{}}Ground-truth = ``B-ORG"\\ Same mention annotation = ``B-ORG"\\ \end{tabular}} & ``B-ORG" & ``I-ORG" & ``B-PER"& ``I-PER" & ``B-LOC"& ``I-LOC"& ``B-MISC"& ``I-MISC" & ``O"  \\ 
\cline{2-10}
                                                                                                                                & $\bm{\beta}_1$     & $\bm{\beta}_2$     & $\bm{\beta}_3$     & $\bm{\beta}_4$    & $\bm{\beta}_5$     & $\bm{\beta}_6$     & $\bm{\beta}_7$      & $\bm{\beta}_8$      & $\bm{\beta}_9$  \\
\hline
\end{tabular}
\end{table*}

\begin{table}
\centering
\caption{BC2GM: the selected vector of $\bm{\alpha}$ parameter matrix }
\begin{tabular}{|c|c|c|c|} 
\hline
\multirow{2}{*}{\begin{tabular}[c]{@{}c@{}}Ground-truth = ``I-GENE"\\Previous annotation = ``B-GENE"\\ \end{tabular}} & ``B-GENE" & ``I-GENE" & ``O" \\ 
\cline{2-4}
                                                                                                                                  & $\bm{\alpha}_1$     & $\bm{\alpha}_2$       & $\bm{\alpha}_3$  \\
\hline
\end{tabular}
\end{table}

\begin{table}
\centering
\caption{BC2GM: the selected vector of $\bm{\beta}$ parameter matrix}
\begin{tabular}{|c|c|c|c|} 
\hline
\multirow{2}{*}{\begin{tabular}[c]{@{}c@{}}Ground-truth = ``B-GENE"\\Previous annotation = ``B-GENE"\\ \end{tabular}} & ``B-GENE" & ``I-GENE" & ``O" \\ 
\cline{2-4}
                                                                                                                                  & $\bm{\beta}_1$     & $\bm{\beta}_2$       & $\bm{\beta}_3$  \\
\hline
\end{tabular}
\end{table}

We can see from Figure 11 and 12 that by aggregating crowd annotations the probability of including the ground-truth label for model learning can be improved. Furthermore, annotators have limited choices with the smaller label set and then the obtained crowd annotations contain much less wrong and invalid label information, which greatly help improve the performance of SA-SLC.

\begin{figure}
\centering
\includegraphics[width=3.7in,height = 3.5in ]{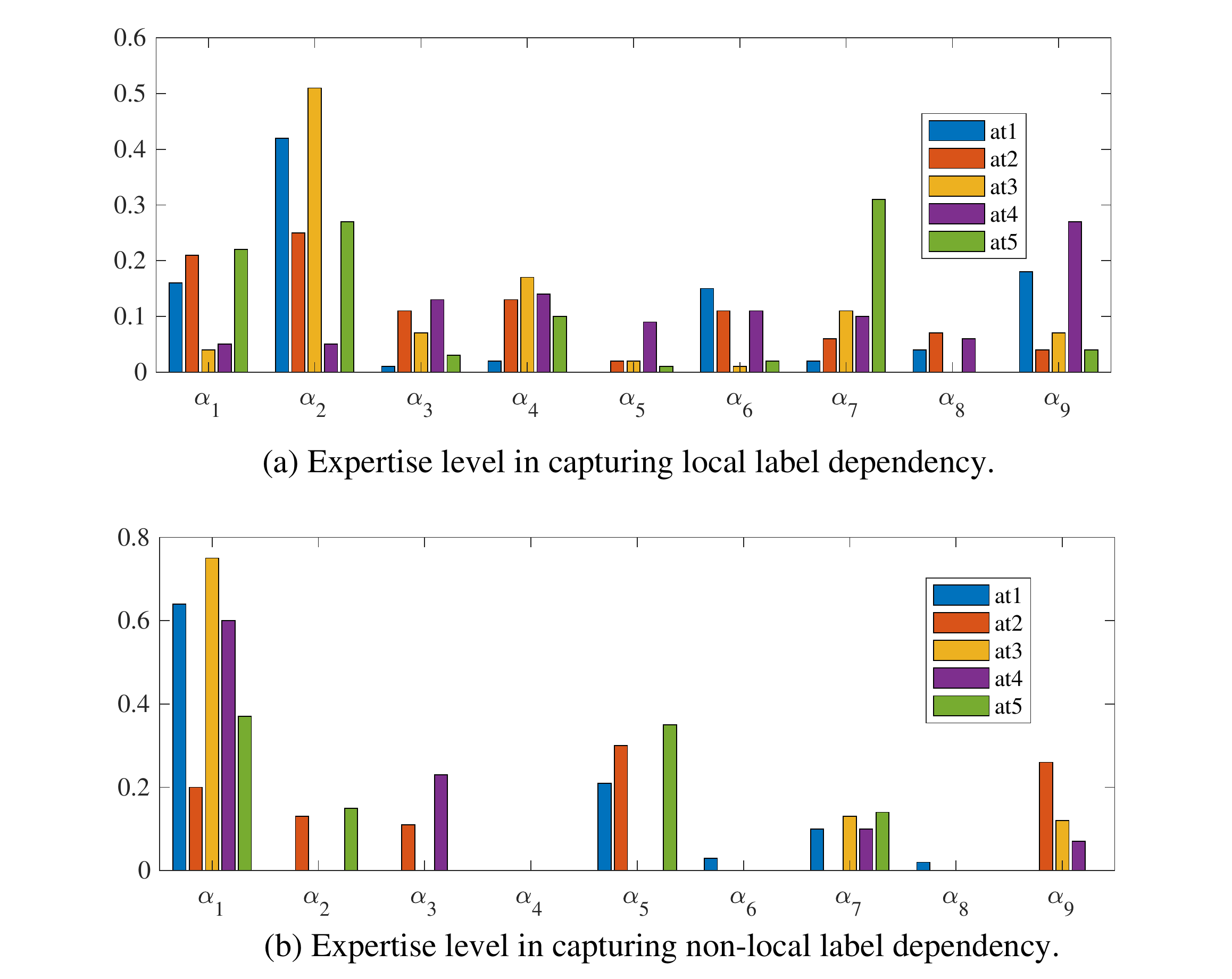}
\caption {CoNLL2003 dataset: expertise level in capturing local and non-local label dependency of annotators ($p = 0.5$).}
\label{fig:secondfigure}
\end{figure}

\begin{figure}
\centering
\includegraphics[width=3.7in,height = 3.5in ]{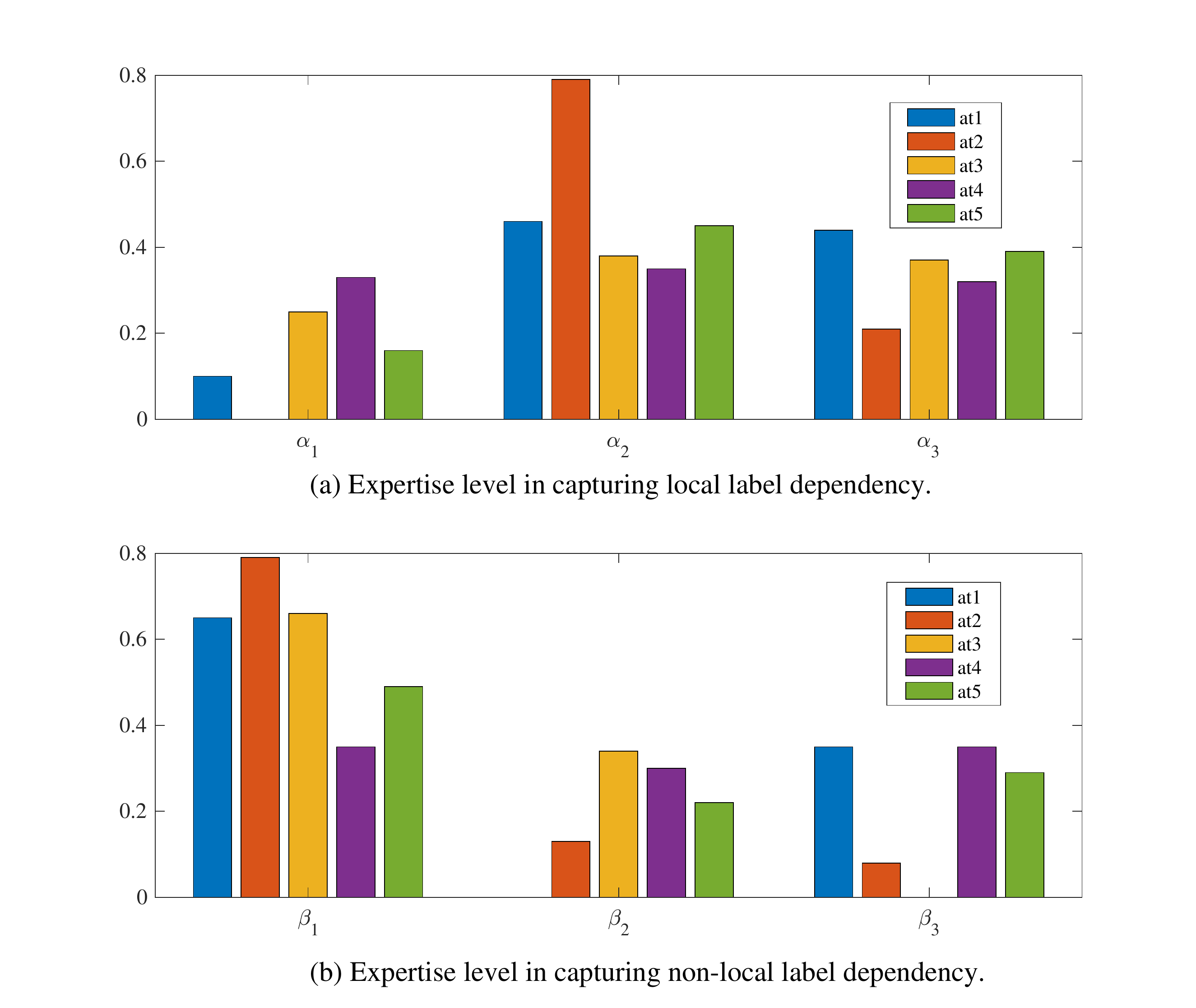}
\caption {BC2GM dataset: expertise level in capturing local and non-local label dependency of annotators ($p = 0.5$).}
\label{fig:secondfigure}
\end{figure}

Generally, the performance of crowdsourced learning methods is significantly correlated with annotators' expertise. Aggregating the crowd annotations or inferring the ground-truth label from crowd annotations is less likely to achieve good results with lower label quality of crowds (i.e. lower $p$). Different from the wrapper mothods (e.g. MVtoken and DS) that infers the ground-truth label for the subsequent classifier learning, SA-SLC, Sembler and HMM-Crowd estimate the label quality and build the classifier simultaneously, which greatly reduces the negative effect of unreliable annotation. Compared with Sembler and HMM-Crowd, SA-SLC improves the estimation of the quality of label sequence by considering annotator's expertise in capturing local and non-local label dependency. Furthermore, SA-SLC incorporates rich and valid label information by exploring label consistency among annotators, which helps improve the performance of learning from crowd annotations with much lower $p$.

\begin{table*}[h]
\centering
\caption{The performance of comparing methods with different number of annotators on CoNLL 2003 and BC2GM datasets (\%).}
\begin{tabular}{cl|ccccc|ccccc} 
\hline
\multirow{2}{*}{\#annotators} & \multicolumn{1}{c|}{\multirow{2}{*}{Method}} & \multicolumn{5}{c|}{CoNLL 2003}                                                          & \multicolumn{5}{c}{BC2GM}                                                                                                                  \\
                             & \multicolumn{1}{c|}{}                        & 0.1             & 0.3             & 0.5             & 0.7             & 0.9             & 0.1                      & 0.3                       & 0.5                       & 0.7                       & 0.9                         \\ 
\hline
\multirow{6}{*}{K = 5}       & MVtoken                                      & 35.75           & 49.20           & 64.34           & \textbf{67.97}  & 67.92           & 48.38                    & 45.94                     & 46.77                     & 54.12                     & 58.00                       \\
                             & DS                                           & 31.67           & \textbf{51.71}  & \textbf{65.53}  & 67.62           & 67.96           & 42.83                    & 44.96                     & 50.11                     & 58.24                     & 54.56                       \\
                             & MACE                                         & 24.01           & 40.93           & 46.15           & 53.73           & \textbf{68.68}  & 30.49                    & 50.26 & 59.82 &  55.17 & 61.86  \\
                             & Sembler                                      & 36.08           & 44.00           & 51.22           & 61.02           & 68.45           & 47.00                    & 48.74                     & 60.24                     & 59.25                     & 65.20                       \\
                             & HMM-Crowd                                     & 37.07           & 43.72           & 54.12           & 62.35           & 67.80           & 42.05                    & 48.03                     & 59.90                     & 59.95                     & 62.38                       \\
                             & SA-SLC                                       & \textbf{40.08}  & 46.91           & 60.21           & 63.56           & 67.47           & \textbf{50.00}           & \textbf{50.14}            & \textbf{62.29}            & \textbf{60.10}            & \textbf{67.26}              \\ 
\hline
\multirow{6}{*}{K = 10}      & MVtoken                                      & 32.27           & 63.61           & \textbf{67.21}  & 68.25  & 67.77           & 47.01                    & 58.47                     & 66.86                     & \textbf{72.57}            & \textbf{72.05}              \\
                             & DS                                           & 38.25           & \textbf{65.67}  & 66.59           & 68.00           & 67.82           & 43.76                    & 48.79                     & 57.40                     & 57.25                     & 56.73                       \\
                             & MACE                                         & 25.10           & 45.09           & 52.75           & 55.23           & 63.17           &38.91 & 52.81                     & 61.74                     & 52.18                     & 56.79                       \\
                             & Sembler                                      & 40.78           & 47.28           & 56.20           & 66.31           & 67.01           & 48.95                    & 55.71                     & 61.46                     & 65.32                     & 71.04                       \\
                             & HMM-Crowd                                     & 41.50           & 47.29           & 60.07           & \textbf{68.30}           & 68.10          & 45.50                    & 49.37                     & 55.39                     & 60.09                     & 63.97                       \\
                             & SA-SLC                                       & \textbf{43.04}  & 48.32           & 61.55           & 66.92           & \textbf{68.21}  & \textbf{52.31}           & \textbf{58.50}            & \textbf{66.27}            & 69.70                     & 71.31                       \\ 
\hline
\multirow{6}{*}{K = 15}        & MVtoken                                      & 43.92           & 66.09           & \textbf{68.04}  & 67.78           & 67.81           & 51.63                    & 61.86                     & \textbf{71.49}            & 73.22                     & 73.25                       \\
                             & DS                                           & 46.24           & \textbf{66.13}  & 67.70           & \textbf{67.92}  & 67.82           & 53.07                    & 46.74                     & 58.30                     & 50.70                     & 56.69                       \\
                             & MACE                                         & 30.00           & 48.85           & 57.21           & 60.25           & 65.36           & 44.56                    & 57.58                     & 62.36                     & 56.10                     & 56.72                       \\
                             & Sembler                                      & 40.09           & 52.32           & 60.85           & 65.71           & 68.91           & 52.45                    & 61.30                     & 69.24                     & 72.35                     & 74.23                       \\
                             & HMM-Crowd                                     & 45.74           & 55.65           & 62.59           & 66.75           & 67.03           & 50.01                    & 57.40                     & 68.32                     & 73.57                     & 71.02                       \\
                             & SA-SLC                                       & \textbf{46.80}  & 57.00           & 62.54           & 65.38           & \textbf{70.02}  & \textbf{55.19}           & \textbf{62.49}            & 70.05                     & \textbf{75.30}            & \textbf{77.26}              \\
\hline
\end{tabular}
\end{table*}

We further investigate the influence of number of annotators. By increasing K to 10 and 15, we record the F1 score of each comparing method and present the results in Table \uppercase\expandafter{\romannumeral8} and \uppercase\expandafter{\romannumeral9}. For most of comparing methods, increasing the number of annotators can help improve the performance. For example, the performance of SA-SLC on CoNLL 2003 dataset ($p = 0.1$) improves by 7.40\% and 16.77\% with $K = 10$ and $K = 15$ respectively. In most cases the performance of crowdsourced learning methods is positively correlated with the expertise level with specified $K$. For example, the average increase rate over varying $p$ of SA-SLC and HMM-Crowd in BC2GM dataset with $K = 10$ is $8.15\%$ and $8.91\%$ respectively. Furthermore, SA-SLC always outperforms the other joint models (i.e. Sembler and HMM-Crowd) by varying $K$. It should be noted that in some cases MVtoken achieves competitive results compared with SA-SLC. For the models that directly infer the ground-truth label from crowd annotations, increasing $K$ relaxes the limitations of label space for inference, which may improve the quality of the inferred ground-truth label.

\begin{table*}[h]
\centering
\caption{The performance of comparing methods with different number of annotators on NCBI-disease and JNLPBA datasets (\%).}
\begin{tabular}{cl|ccccc|ccccc} 
\hline
\multirow{2}{*}{\#annotators} & \multicolumn{1}{c|}{\multirow{2}{*}{Method}} & \multicolumn{5}{c|}{NCBI-disease}                                                                                         & \multicolumn{5}{c}{JNLPBA}                                                               \\
                              & \multicolumn{1}{c|}{}                        & 0.1                        & 0.3             & 0.5                        & 0.7                         & 0.9             & 0.1             & 0.3             & 0.5             & 0.7             & 0.9              \\ 
\hline
\multirow{6}{*}{K = 5}        & MVtoken                                      & 60.71                      & 64.73           & 64.83                      & 61.51                       & 69.63           & 34.42           & 41.49           & 52.22           & \textbf{60.64}  & \textbf{64.14}   \\
                              & DS                                           & 58.24                      & 60.90           & 69.19                      & 69.68                       & 71.33           & 16.44           & 27.92           & 41.92           & 49.52           & 50.50            \\
                              & MACE                                         & 44.69                      & 51.86           & 62.70 & 67.91                       & 69.60           & 28.50           & 40.52           & 46.62           & 51.71           & 56.70            \\
                              & Sembler                                      & 56.11                      & 61.59           & 67.60                      & 68.94                       & 65.90           & 32.73           & 40.97           & 43.20           & 56.27           & 55.32            \\
                              & HMM-Crowd                                     & 55.90                      & 61.25           & 62.79                      & 68.65                       & 70.25           & 29.54           & 41.08           & 47.27           & 55.29           & 62.31            \\
                              & SA-SLC                                       & \textbf{60.94}             & \textbf{65.17}  & \textbf{70.25}             & \textbf{70.50}              & \textbf{75.27}  & \textbf{37.62}  & \textbf{42.01}  & \textbf{52.80}  & 58.30           & 58.37            \\ 
\hline
\multirow{6}{*}{K = 10}       & MVtoken                                      & 61.30                      & 65.98           & 67.17                      & 70.81                       & 72.47           & 31.70           & \textbf{50.69}  & \textbf{60.00}  & \textbf{62.77}  & 62.58            \\
                              & DS                                           & 58.79                      & 67.36           & \textbf{69.83}             & 66.61                       & \textbf{73.50}  & 25.24           & 38.80           & 44.94           & 47.09           & 48.31            \\
                              & MACE                                         &  45.51 & 58.00           & 62.09                      & 69.43& 73.17           & 31.76           & 48.20           & 52.37           & 60.75           & 62.25            \\
                              & Sembler                                      & 60.23                      & 66.26           & 69.27                      & 71.55                       & 71.68           & 35.00           & 45.49           & 53.28           & 58.31           & 61.77            \\
                              & HMMcrowd                                     & 62.45                      & 67.24           & 69.45                      & 70.52                       & 71.09           & 35.21           & 48.69           & 53.74           & 58.89           & 62.20            \\
                              & SA-SLC                                       & \textbf{64.70}             & \textbf{69.38}  & 69.71                      & \textbf{72.02}              & 72.35           & \textbf{38.50}  & 48.57           & 55.90           & 59.24           & \textbf{62.61}   \\ 
\hline
\multirow{6}{*}{K = 15}         & MVtoken                                      & 63.66                      & 69.69           & 70.17                      & 71.55                       & \textbf{73.43}  & 37.05           & \textbf{56.45}  & \textbf{63.53}  & 62.90           & 63.17            \\
                              & DS                                           & 61.05                      & 67.73           & \textbf{70.62}             & 68.52                       & \textbf{73.43}  & 32.27           & 42.19           & 44.23           & 45.61           & 47.17            \\
                              & MACE                                         & 50.02                      & 58.47           & 68.36                      & 71.08                       & 72.56           & 32.50           & 52.71           & 53.89           & 60.73           & 63.09            \\
                              & Sembler                                      & 63.30                      & 68.25           & 69.83                      & \textbf{72.26}              & 72.50           & 37.27           & 50.92           & 57.25           & 59.10           & 62.07            \\
                              & HMM-Crowd                                     & 63.68                      & 69.27           & 69.74                      & 72.01                       & 71.90           & 37.08           & 52.39           & 57.10           & 60.24           & 61.05            \\
                              & SA-SLC                                       & \textbf{66.20}             & \textbf{70.31}  & 70.24                      & 71.38                       & 72.74           & \textbf{39.15}  & 52.76           & 58.31           & \textbf{63.69}  & \textbf{64.52}   \\
\hline
\end{tabular}
\end{table*}

\subsection{Real-world application in Amazon’s Mechanical Turk}

Rodrigues et al. \cite{re20} selected 400 news articles from CoNLL 2003 shared NER task and put them on Amazon’s Mechanical Turk (AMT) to collect crowd labels.There are total 47 annotators and the average number of annotators per article is 4.9. In this paper, after pre-processing these crowd-labeled data we select 3000 sentence-level sequences for training, and use CoNLL 2003 development and test data. Table \uppercase\expandafter{\romannumeral10} shows the results of predicting CoNLL 2003 test data.We can see that joint models significantly outperforms the wrapper methods and SA-SLC achieves the highest recall and F1 score. Compared to the results of Task 2 reported by Nguyen et al. \cite{re22}, HMM-Crowd then CRF (or LSTM) achieves better results than the joint model HMM-Crowd. Training the powerful sequence labeling model (e.g. CRF and LSTM) on the aggregated labels obtained from HMM-Crowd may improve the performance.

\begin{table}[H]
\centering
\caption{Performance of comparing methods learned from AMT CoNLL 2003 (\%).}
\begin{tabular}{lccc} 
\hline
\multicolumn{1}{c}{Method} & Precision & Recall & F1 score  \\ 
\hline
MVtoken                    & 67.80     & 46.51  & 51.01     \\
DS                         & 67.00     & 47.73  & 53.29     \\
MACE                       & 63.32     & 45.72  & 51.32     \\
Sembler                    & 71.02     & 58.79  & 62.47     \\
HMM-Crowd                  & 69.82     & 57.17  & 61.42     \\
SA-SLC                     & \textbf{72.71}    & \textbf{60.02}  & \textbf{64.14}     \\
\hline
\end{tabular}
\end{table}

In NER task, capturing internal label dependency is important for an annotator to provide consistent and reliable annotations. SA-SLC effectively utilize two confusion matrices to measure annotator's expertise in capturing local and non-local label dependency. Table \uppercase\expandafter{\romannumeral11} presents a sentence with crowd annotation from AMT. It seems that the third annotator provides consistent annotation but two ``Shanghai" are expected to be labeled as named entity ``B-LOC". This wrong label assignment can be accurately characterized by the two confusion matrices of SA-SLC that reflect the annotator's reliability in assigning label ``B-LOC" and considering non-local label dependency.

\begin{table*}
\centering
\caption{A crowd-annotated example of CoNLL 2003 shared NER task}
\scalebox{0.85}{
\begin{tabular}{ccccccccccccccccccc} 
\hline
Sentence    & Traders & in & Shanghai & said & on & Thursday & they & were & unaware & of & movements & out & of & the & Shanghai & bonded & warehouses & .  \\ 
\hline
annotator 1 & O       & O  & B-LOC    & O    & O  & O        & O    & O    & O       & O  & O         & O   & O  & O   & B-LOC~   & O~     & O          & O  \\
annotator 2 & O       & O  & B-LOC    & O    & O  & O        & O    & O    & O       & O  & O         & O   & O  & O   & B-LOC    & O      & O          & O  \\
annotator 3 & O       & O  & O        & O    & O  & O        & O    & O    & O       & O  & O         & O   & O  & O   & O        & O      & O          & O  \\
\hline
\end{tabular}}
\end{table*}

\section{Conclusion}
In this paper, we propose SA-SLC to model crowd sequential annotations for sequence labeling. The proposed SA-SLC builds a conditional probabilistic model to jointly model sequential data and annotators' expertise. Through modeling the expert level of the annotator in capturing internal label dependency for sequential annotation, SA-SLC improves the estimation of the quality of label sequence from crowd annotations, which greatly reduces the negative effect of unreliable annotations in the optimization. Furthermore, a valid label sequence inference method is designed to accelerate the marginalization of SA-SLC, which further improves the quality of possible ground-truth label sequences. We conducted the experiments on four sequential datasets with synthetic crowd annotations. By varying expertise level and the number of annotators, in most cases SA-SLC performs better than the other comparing methods. In the future, we will consider more appropriate (or complex) annotator's behavior in sequential annotation.

\section*{Acknowledgment}
This work was supported by the Hong Kong Research Grants Council (RGC) General Research Fund (GRF) Project under Grant 9042996.
\bibliographystyle{ieeetr}
\bibliography{ref}

\end{document}